\pdfoutput=1
\documentclass[review,times,10pt]{elsarticle}

\usepackage{amssymb}
\usepackage{amsmath}
\usepackage{multirow} 
\usepackage{booktabs}
\usepackage{graphicx}
\usepackage{colortbl}
\usepackage{pifont}

\usepackage{subcaption}
\usepackage{color}
\usepackage{rotating}

\journal{Patten Recognition}

\begin{document}

\begin{frontmatter}

\title{Spiking Pyramid Wavelet Transformation for High-efficient and Low-energy Image Restoration}

\author[NJU]{Chen Zhao}  
\author[NJUST]{Xiantao~Hu}  
\author[CM]{Song~Wu \corref{cor}}  
\author[CM]{Qian~Wang} 
\author[USTC]{Chen~Wu}
\author[NJU]{Rui Xie}
\author[NJU]{Jian~Yang}
\author[NJU]{Ying~Tai \corref{cor}}

\affiliation[NJU]{organization={School of Intelligence Science and Technology},
            addressline={Nanjing University}, 
            city={Suzhou},
            postcode={215163}, 
            state={Jiangsu},
            country={China}}
            
\affiliation[NJUST]{organization={PCA Lab, Key Lab of Intelligent Perception and Systems for High-Dimensional Information of Ministry of Education, School of Computer Science and Engineering},
            addressline={Nanjing University of Science and Technology}, 
            city={Nanjing},
            postcode={210094}, 
            state={Jiangsu},
            country={China}}
\affiliation[USTC]{organization={University of Science and Technology of China},
            city={Hefei},
            postcode={230026}, 
            state={Anhui},
            country={China}}
\affiliation[CM]{organization={China Mobile Institute},
            city={Beijing},
            postcode={100032}, 
            country={China}}

\cortext[cor]{Corresponding authors}

\begin{abstract}

Spiking neural networks (SNNs) have garnered significant interest in computer vision due to their potential for efficiency and biological inspiration. While spiking CNN-based methods have shown promise for image restoration (IR) tasks, their performance is constrained by the inherent receptive field limitations of CNN operations. In the paper, we explore the benefits of discrete wavelet transformation  and propose a spiking pyramid wavelet-based model (SPWM) for high-efficient and low-energy target. Specifically, we develop a spiking dual pyramid wavelet (SDPW) block to model long-range dependency and exploit the properties of the degradation in the wavelet domain.  Experimental results on several benchmarks demonstrate that SPWM significantly lowers computational costs and energy consumption while maintaining image quality. Our method showcases the potential of SNNs in the field of IR, offering new insights for future applications of resource-limited devices.

\end{abstract}

\end{frontmatter}




\section{Introduction}

Image restoration (IR) aims to reconstruct high-quality images from degraded low-quality inputs, serving a crucial role in numerous real-world applications \cite{,jiang2025learning,wang2025learning,liu2019griddehazenet,zhao2024cycle,tai2026addsr}. It encompasses a diverse array of sub-tasks, such as deraining, dehazing and low light enhancement. The inherent challenge of IR stems from its ill-posed nature, 
\textit{i.e.}, the irreversible process of image degradations. 
These degradations not only impair the visual quality of images but also impose significant constraints on various vision tasks. Recently, deep learning-based IR methods have achieved satisfactory results in various restoration tasks, such as deraining, dehazing, deblurring, low-light enhancement, etc.

Existing deep learning-based methods have achieved promising results for IR tasks, primarily including convolutional neural network (CNN)-based models \cite{tai2017image} and transformer based models \cite{liang2021swinir,zhao2026learning,zeng2025semantic,zhao2025multi}. However, these CNN-based methods usually suffer from the inherent limitation of CNNs, the constrained receptive field. In order to alleviate this problem, numerous works \cite{zamir2021multi} meticulously crafted network architectures, significantly increasing model complexity and impacting efficiency. Although transformers can achieve global interaction modeling, they face a significant problem as the self-attention mechanism results in quadratic space and time complexity with the number of tokens. Existing works \cite{chen2023learning} have been proposed to reduce model parameters or complexity (FLOPs) via token sparsification and  pruning. However, achieving a better trade-off between performance and efficiency remains a significant challenge.

\begin{table}
{
	\centering
	\caption{Attempt for spiking transformer on the Rain200H. }
	
	\resizebox{1.0\width}{!}{
		\begin{tabular}{ccccc}
			\toprule
			\multirow{1}{*}{Model} & \multirow{1}{*}{FLOPs (G)$\downarrow$} & \multirow{1}{*}{Params(M)$\downarrow$} & \multirow{1}{*}{PSNR$\uparrow$} & \multirow{1}{*}{SSIM$\uparrow$} \\ \hline
			ESDNet                       & \textbf{7.320  }                          &\textbf{0.165  }                                  & 29.50                        & 0.9079                        \\
			
			ESDformer                       &  89.16                      &2.08                                     & \textbf{29.76}                        & \textbf{0.9100 }                        \\
			\bottomrule
		\end{tabular}
	}
	
	
	\label{tab1} }
\end{table}
Spiking neural networks (SNNs) have attracted considerable research interest in the field of computer vision \cite{yao2024spike,su2023deep}. 
SNNs, known as the third generation of neural networks 
~\cite{roy2019towards} potentially serves as a more efficient and biologically inspired way for various vision tasks, with low computational complexity and energy consumption, which are particularly well-suited for deployment on edge devices with limited resources. 
Recent Spiking CNN-based IR methods \cite{song2024learning} have emerged, but their effectiveness is hindered by the constrained local receptive fields characteristic of CNNs. 
This inspires us to explore spiking transformers \cite{yao2024spike} to alleviate the limitations of spiking CNNs.
Table~\ref{tab1} shows that spiking transformer~\cite{yao2024spike} outperforms spiking CNN-based method~\cite{song2024learning}, where ESDformer represents the approach of replacing the spiking residual block in ESDNet \cite{song2024learning} with a spiking transformer \cite{yao2024spike}.
However, directly using spiking transformers results in a significant FLOPs and Params cost, which conflicts with our goal of high efficiency. Thus, it is crucial to explore a spiking operator that \textit{balances efficiency with long-range dependency modeling}.

\begin{figure*}[!t]
	\centering
	\includegraphics[width=1 \linewidth]{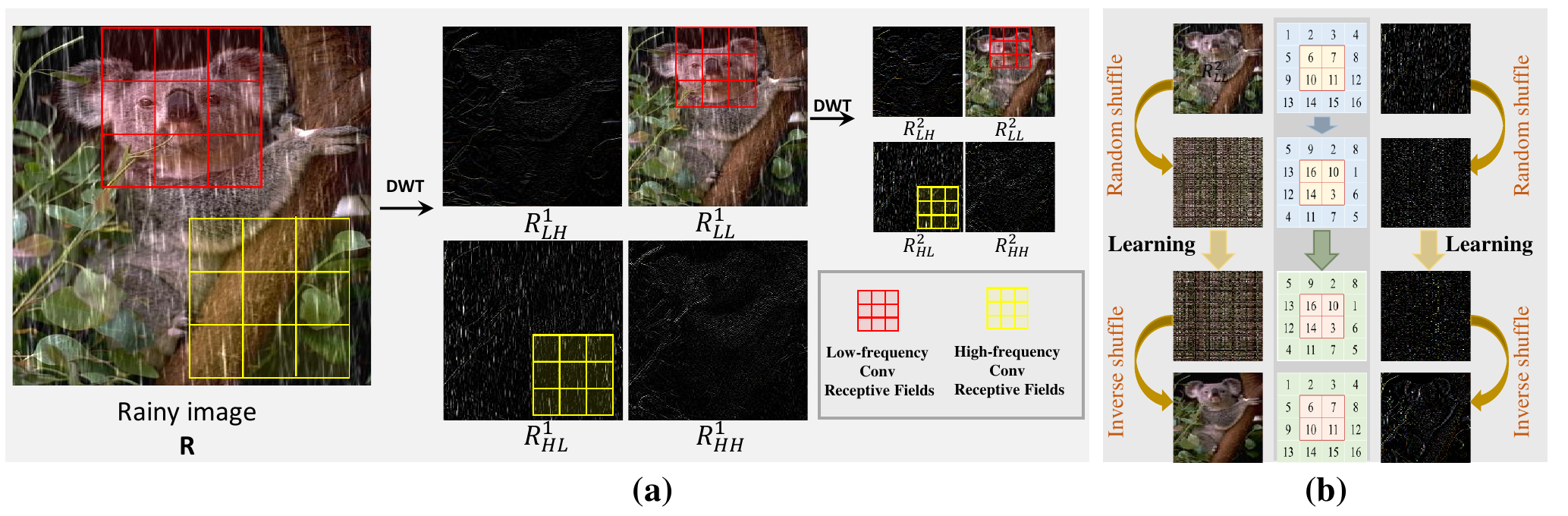}
	\caption{Our motivations. The degraded images can be decomposed into low-frequency sub-image ($LL$) and high-frequency sub-images ($LH, HL, HH$) using discrete wavelet transformations (DWT). We use multi-level DWT to explore the properties of degraded images, as illustrated in (a). To further achieve long-range dependency modeling, we employ a random shuffling (RS) operation, as depicted in (b).
	}
	\label{fig1}
\end{figure*}

\begin{figure}[!t]
	\centering
	\includegraphics[width=1.0\linewidth]{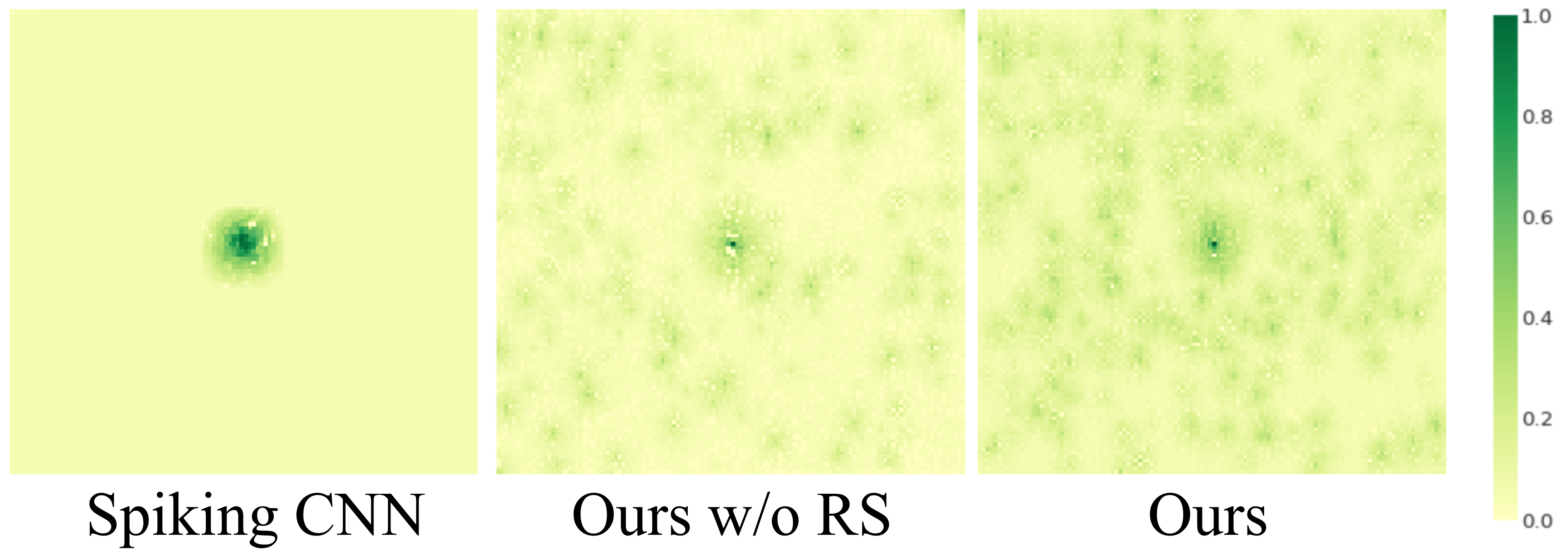}
	\caption{The effective receptive field visualization.}
	\label{fige}
\end{figure}

\begin{table}
	\centering
	\caption{Comparison of metrics between frequency domain components of the degraded image in the wavelet domain and their corresponding components in the ground truth on the Rain200H dataset. 
	}
	\resizebox{0.90\width}{!}{
		
		\begin{tabular}{ccccc}
			
			\toprule
			
			\multirow{1}{*}{PSNR/SSIM
			} & \multirow{1}{*}{LL} & \multirow{1}{*}{LH} & \multirow{1}{*}{HL} & \multirow{1}{*}{HH} \\ \hline
			1-level DWT                       & \textbf{ 12.05}/0.5345
			&    28.69/0.7816
			&       19.96/\textbf{0.3418}
			&   25.87/0.6369
			\\
			
			2-level DWT                       &  \textbf{13.07}/0.5786      &27.63/0.7593                                 & 18.55/\textbf{0.3002}                                         & 23.56/0.5579
			\\
			\bottomrule
		\end{tabular}
	}

	\label{tab2}
\end{table}
Fig.~\ref{fig1} illustrates our motivations schematically. We explore the properties of the degraded images in the Wavelet domain. Specifically, the degraded images can be decomposed into low-frequency sub-image ($LL$) and high-frequency sub-images ($LH, HL, HH$) using discrete wavelet transformations (DWT). From visual results of Fig.~\ref{fig1}~(a), we can find two primary advantages of the wavelet transform: \textit{(1) as the level of wavelet domain increases, it allows for an expanded receptive field, and (2) it may exploit the properties of the degraded images, the rain streaks (degradation) is mainly concentrated in  specific frequency domain.} 
Table \ref{tab2} presents similar results that the PSNR is lowest in the LL frequency domain component, and the SSIM is lowest in the HL frequency domain component, illustrating that the degradation is primarily concentrated in specific frequency components. 
To further achieve long-range dependency modeling, we employ a random shuffling (RS) operation in the wavelet domain. 
Fig.~\ref{fig1}~(b) provides a visualization of this operation.  
Specifically, by randomly shuffling across the spatial domain, every pixel can be placed in any position with equal likelihood. By breaking the original ordered local structure of the features, the RS enables the Conv layer to learn long-range dependencies across different regions of the feature sequence. 
Moreover, the degradation is distributed in a homogenized space, which can reduce the difficulty of restoration~\cite{xiao2024homoformer}.  Fig.~\ref{fige} illustrates the visualization of the effective receptive field \cite{luo2016understanding}, demonstrating that our method effectively enables long-range dependency modeling.
It's worth noting that DWT and RS operations are \textit{cost-effective}, without 
any \textit{additional parameters or FLOPs}.

In this paper, we propose a spiking pyramid wavelet-based model, called SPWM, to achieve our high-efficient and low-energy target. Specifically, 
to model long-range dependency and exploit the properties of the degradation in the wavelet domain, we propose a spiking dual pyramid wavelet (SDPW) block, which consists of a spiking pyramid wavelet unit (SPWU) and a spiking pyramid shuffle unit (SPSU). SDPW can convert the input into spike signals, and then we introduce multi-dimensional attention (MDA) \cite{yao2023attention}  to adaptively adjust spike responses, alleviating information missing due to binary activation of spike neuron. 
Moreover,  we apply a multi-scale progressive fusion strategy and design an ultra-lightweight spatial-channel adjustment module (SCAM) to  mitigate the information loss in converting from discrete pulse sequences to continuous pixel values.

In summary, the main contributions are as follows:

\begin{itemize}
	\item We explore the benefits of wavelet transform and random shuffle operations and propose a spiking pyramid wavelet-based model (SPWM) for high-efficient and low-energy  IR, which provides a novel perspectives for applications of resource-limited devices. 
	\item We propose a spiking dual pyramid wavelet (SDPW) block to model long-range dependency and exploit the properties of the degradation in the wavelet domain.
	\item Experimental results compared with ANNs-based methods on several IR datasets demonstrate that our SPWM can obtain
	comparable restoration performance. 
\end{itemize}

\begin{figure*}
	\centering
	\includegraphics[width=1 \linewidth]{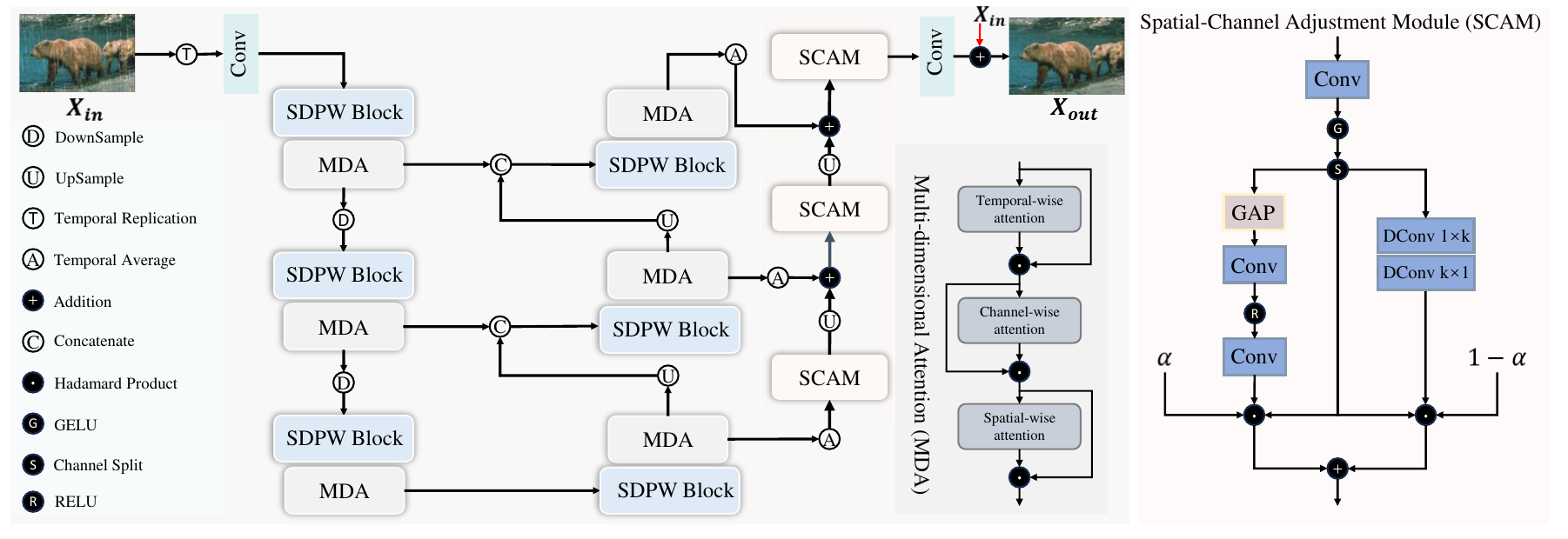}
	\caption{Overall  architecture of our proposed spiking pyramid wavelet-based model (SPWM). The core components are the proposed spiking dual pyramid wavelet (SDPW) block, multi-dimensional attention (MDA), and the designed spatial-channel adjustment module (SCAM). 
	}
	\label{fig2}
\end{figure*}

\vspace{-3mm}
\section{Related Work}
\subsection{Image Restoration}
Deep learning-based methods have achieved outstanding performance for various computer vision tasks, including image restoration (IR) tasks \cite{dong2024ecmamba,zhao2024toward,zhou2024glare,zhou2025low,zhao2025spectral,lu2025does,dong2025mamba,zhao2025learning}. Deep learning-based IR methods are primarily categorized into CNN-based and Transformer-based models.  SRCNN \cite{dong2015image} was introduced for image super-resolution, while   DnCNN \cite{zhang2017beyond} was developed for denoising, and PReNet \cite{ren2019progressive} was created for deraining. These foundational works sparked a surge of CNN-based techniques, enhancing representation capabilities of deep neural networks for IR tasks. For the Transformer, SwinIR \cite{liang2021swinir} and Restormer \cite{zamir2022restormer} were introduced  for various restoration tasks, leading to numerous Transformer-based approaches \cite{xiao2024homoformer}. However, it is difficult for deep learning-based approaches to achieve a better trade-off between performance and efficiency. 
To address this issue, we propose a novel perspective, leveraging Spiking Neural Networks (SNNs) to achieve high-efficiency and low-energy restoration.

\vspace{-3mm}
\subsection{Spiking Neural Networks}
Recently, SNNs have made significant strides in the field of computer vision. Utilizing surrogate gradient functions for direct training of SNNs has significantly reduced the simulation time step while achieving satisfactory results \cite{wang2023masked}. EMS-YOLO \cite{su2023deep} achieved performance equivalent to an ANN with the same SNNs architecture using only 4 time steps for object detection. \cite{zhouspikformer} introduced spiking transformer and achieved good performance for classification tasks. Spiking-driven transformer \cite{yao2024spike} is proposed to achieve  good performance on many vision tasks including classification, segmentation and detection. For IR tasks, spiking CNN-based methods \cite{song2024learning} were proposed, but their performance is limited by the inherent limitations of CNN operations, preventing it from achieving better results. However, the direct employment of spiking transformers \cite{yao2024spike,zhao2025zero,zhouspikformer} contradicts our high-efficiency goal. Therefore, we explored spiking operators based on frequency decoupling for high-efficiency goal.

\vspace{-3mm}
\section{Methodology}
\subsection{Overall Network Architecture}

Given a single degraded image as input $X_{in}$, our goal is to learn a spiking neural network to generate an high-quality output $X_{out}$ that eliminates various degradations.
The overall architecture of our SPWM is shown in Fig.~\ref{fig2}, which can be broadly divided into three stages: temporal sequence encoding, spiking transformation, and feature conversion. 
Firstly, to adapt to the spatiotemporal properties of SNNs, we employ a direct encoding strategy where the static input $X_{in} \in \mathbb{R}^{H \times W \times C}$ is repeated across all time steps to generate a consistent sequence $\mathcal{X} = \{X_t\}_{t=1}^{T}$ with $X_t = X_{in}$. This constant input current is then integrated by the initial spiking neurons to convert pixel intensities into discrete binary spike trains, effectively leveraging the network's internal membrane potential dynamics without the stochastic information loss of traditional Poisson encoding.
Subsequently, we employ a convolution layer to extract features from the $\mathbf{X}$.  The captured features will be fed into our proposed spiking dual pyramid wavelet (SDPW) block to conduct spike transformation, aiming to further learn complex degraded feature representations. In SDPW block, we use the Leaky Integrate-and-Fire neurons (LIF) spike neuron layer, which can convert and weight the inputs to generate output spike sequences. The value of the output spike from LIF is 1 when the membrane potential exceeds the threshold, otherwise 0. 
Moreover, we introduce multi-dimensional Attention (MDA) \cite{yao2023attention}  to adaptively adjust spike responses by applying attention
weights, alleviating information missing due to binary activation of spike neuron. Finally, after several stacked SDPW block and MDA, 
we apply a multi-scale progressive strategy to convert the features extracted from the spike sequence into continuous value representations. Furthermore, we design an ultra-lightweight spatial-channel adjustment module (SCAM) to enhance the reconstructed real value feature.  Finally, the proposed SPWM can generate high-quality output $X_{out}$  via a 3 × 3 convolution layer.

\begin{figure*}
	\centering
	\includegraphics[width=1 \linewidth]{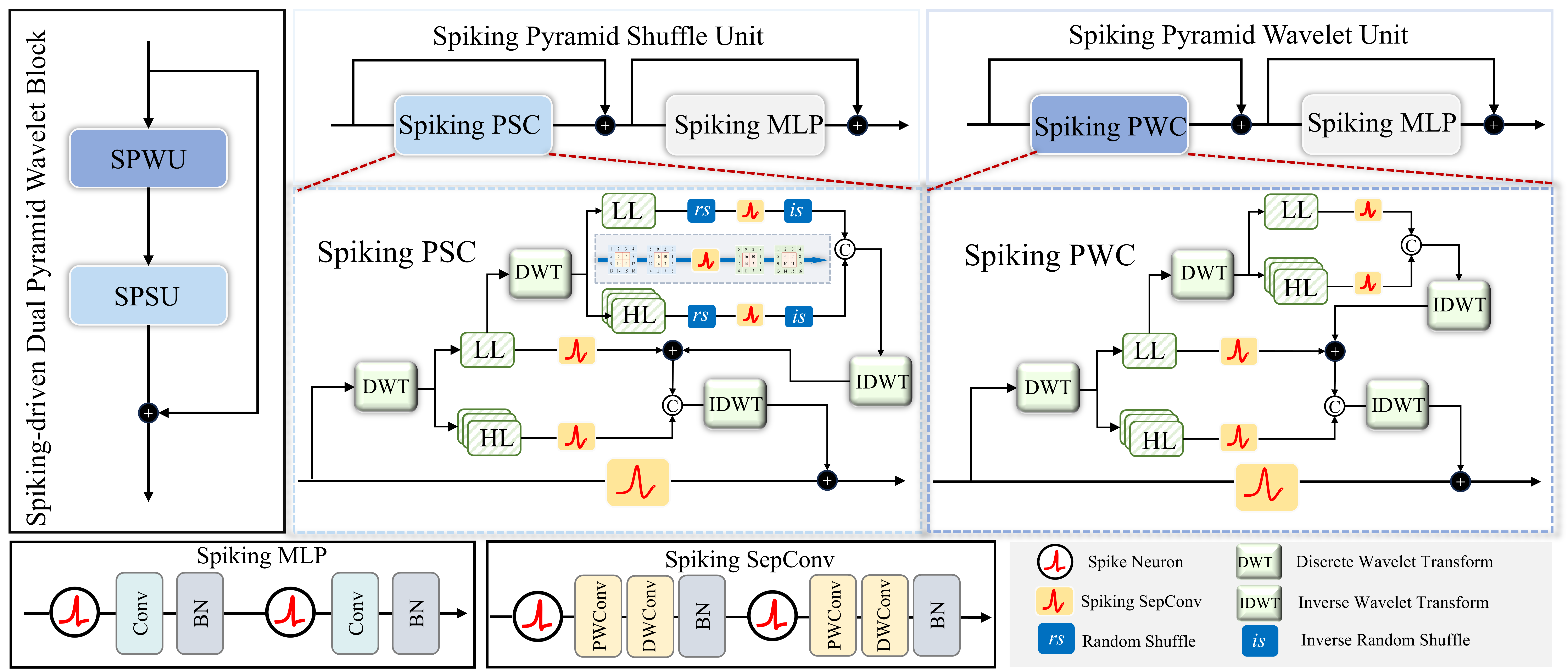}
	\caption{The structure of the proposed SDPW block. The figure displays the network details in the 2-level wavelet domain.}
	\label{figblock}
\end{figure*}

\vspace{-2mm}
\subsection{Discrete Wavelet Transform}
To exploit the characteristics of different frequency signals and achieve decoupling purposes, many works have explored Discrete Wavelet Transform (DWT) for IR tasks \cite{zhao2024wavelet,zhao2026luve,jiang2023low,11513016,zhao2026ultrahr,zhou2026learning}.  Given an input  $X\in\mathbb{R}^{H\times W\times c}$, we use DWT with Haar wavelets to decompose the input. Haar wavelets are efficient and consist of the low-pass filter $L$, and the high-pass filter $H$, as follows:
\begin{equation}L=\frac1{\sqrt{2}}[1,1]^T,H=\frac1{\sqrt{2}}[1,-1]^T.\end{equation}

We can obtain four sub-bands in 1-level DWT, which can be expressed as:
\begin{equation}
	X_{LL},\{X_{LH},X_{HL},X_{HH}\}=\text{DWT}(X),
\end{equation}
where $X_{LL},\{X_{LH},X_{HL},X_{HH} \} \in \mathbb{R}^{\frac H2\times\frac W2\times c}$ represent the low-frequency coefficient of the input and high-frequency coefficients in the vertical, horizontal, and diagonal directions, respectively. In particular, the low-frequency coefficient contains the global information of the original input \cite{jiang2023low}, which can be treated as the downsampled version of the input, and so wavelet-based operators can obtain a bigger receptive field. 
As shown in Fig.~\ref{fig1} (a), the degradation primarily is involved in certain frequency domain components. The metric results in Table ~\ref{tab2} also confirm this. Therefore, wavelet-based operators are better suited to learn the degradation features. To further open the receptive field,  we  perform $N$
times DWT on the low-frequency coefficient:
\begin{equation}
	X_{LL}^{(n)}, \{X_{LH}^{(n)},X_{HL}^{(n)},X_{HH}^{(n)}\}=\mathrm{DWT}(X_{LL}^{(n-1)}),
\end{equation}
where $X_{LL}^{(n)}, \{X_{LH}^{(n)},X_{HL}^{(n)},X_{HH}^{(n)}\}\in\mathbb{R}^{\frac{H}{2^n}\times\frac{W}{2^n}\times c}$, $X_{LL}^{(0)}$ can be seen as the original input $X$. $n$ refers to the current level of DWT. As 
$n$ increases, the receptive field of wavelet-based operators will become larger.

\subsection{Random Shuffling }
Traditional convolutional layers are inherently limited by their local receptive fields, capturing dependencies only within restricted spatial neighborhoods. In the wavelet domain, this locality constraint prevents the model from perceiving global structural correlations across frequency subbands. The RS operation serves as a spatial-frequency remapping mechanism. By randomly permuting the feature sequence, RS effectively ``shuffles'' spatially distant wavelet coefficients into the local window of subsequent convolutions. Formally, let the original sequence be $S = [x_1, x_2, \dots, x_n]$. A local convolution at position $i$ typically only processes elements within its immediate vicinity. After RS, the sequence becomes $S' = [x_{p1}, x_{p2}, \dots, x_{pn}]$ (where $p$ is a random permutation), allowing the same convolution to simultaneously process distant elements (e.g., $x_{16}$ and $x_3$). This mechanism:
\begin{itemize}
    \item Breaks Spatial Locality: It disrupts the fixed grid-based dependency, allowing the convolution to bypass its physical kernel size limits.
    \item Enables Global Interaction: It forces the network to learn correlations between any two regions in the feature map, effectively achieving a ``global'' receptive field with standard convolutional operations.
\end{itemize}

\subsection{Spiking Pyramid Wavelet Transformation}
To model long-range dependency and exploit the properties of the degradation in the wavelet domain, we develop a spiking dual pyramid wavelet (SDPW) block, and the detail structure of SDPW is shown in Fig.~\ref{figblock}. It consists of a spiking pyramid wavelet unit (SPWU) and a spiking pyramid shuffle unit (SPSU). Unlike SPWU, SPSU employs a random shuffle and inverse shuffle operation to model more long-range dependency in the final level of the wavelet domain. Moreover, the degradation is distributed in a homogenized space \cite{xiao2024homoformer}, which can reduce the difficulty of restoration by using our designed SPSU. Both SPWU and SPSU incorporate a structure similar to that of the Meta-Spikeformer \cite{yao2024spike}, which can be described as follows: 
\begin{equation}
	\begin{aligned}X'&=X+SSepConv(X),\\X''&=X'+SMLP(X'),\end{aligned}
\end{equation}
where $SSepConv$ and $SMLP$ refer to spiking separable convolution (SepConv) and spiking Multi-layer Perceptron (MLP), respectively. $SSepConv$ and $SMLP$ can be expressed as:
\begin{equation}
	\begin{aligned}&SSepConv(X)=\mathbf{W}_{d2}\mathbf{W}_{p2}(\mathcal{SN}(\mathbf{W}_{d1}\mathbf{W}_{p1}(\mathcal{SN}(X)))),\\&SMLP(X')=Conv(\mathcal{SN}(Conv(\mathcal{SN}(X')))),\end{aligned}
\end{equation}
where $\mathbf{W}_{p1}$ and $\mathbf{W}_{p2}$ are pointwise convolutions, and $\mathbf{W}_{d1}$ and $\mathbf{W}_{d2}$ are depthwise convolution. $\mathcal{SN}$ is the spike neuron layer. We use the
Leaky Integrate-and-Fire neurons (LIF) \cite{fang2023spikingjelly},  and its dynamic equation can be written mathematically as:
\begin{equation}
	\begin{aligned}
		&M^t=H^{t-1}+X^t, \\
		&S^t=\Theta\left(M^t-V_{th}\right), \\
		&H^t=V_{reset}S^t+\left(\beta M^t\right)\odot\left(\mathbf{1}-S^t\right),
	\end{aligned}
\end{equation}
where $X^t$ is the spatial input at timestep $t$. $M^{t}$ means membrane potential that integrates $X^{t}$ and temporal input $H^{t-1}$.
$\beta$ represents the decay factor and $\odot$ indicates the element-wise multiplication. $\Theta(\cdot)$ denotes a  Heaviside step function which equals 1 for $x\ge 0$ and
0 otherwise. 
When $M^t$ exceeds the firing threshold $V_{th}$, the spiking neuron will fire a spike $S^t$, and temporal output $H^{t}$ is reset to $V_{reset}$. 

Both SPWU and SPSU utilize an enhanced $SSepConv$, with their core components being the spiking pyramid wavelet convolution (spiking PWC) and the spiking pyramid shuffle convolution (spiking PFC), respectively. They can learn complex feature transformations and model long-range dependencies through wavelet transform and random shuffle operations.
{\flushleft\textbf{Spiking PWC.}} 
It alleviates the receptive field limitations of $SSepConv$ through pyramid wavelet framework. Specifically, the spiking PWC learns feature transformations within a multi-level wavelet domain, enhancing the representation capabilities of $SSepConv$. The process of feature transformations in the 1-level wavelet domain can be expressed as:
\begin{equation}
	Y=\mathrm{IDWT}(SSepConv(\mathrm{DWT}(X))),
\end{equation}
where IDWT is the inverse transformation of DWT. This operation not only separates the convolution between the frequency components but also allows a
smaller kernel to operate in a larger area of the original input. Referring to Equation 3, we can perform a similar process in the $n$-level wavelet domain:
\begin{equation}
	Y_{LL}^{(n)},\{{Y}_{LH}^{(n)},{Y}_{HL}^{(n)},{Y}_{HH}^{(n)}\}=SSepConv(\mathrm{DWT}({X}_{LL}^{(n)})).
\end{equation}

To fusion the outputs of the different level frequencies, we utilize the linear operation properties of the DWT and IDWT,  which can be expressed $\mathrm{IWT}(X+Y) = \mathrm{IWT}(X)+\mathrm{IWT}(Y).$ Therefore, performing
\begin{equation}
	Y^{(n)}=\mathrm{IWT}(Y_{LL}^{(n)}+Y_{LL}^{(n+1)},\{{Y}_{LH}^{(n)},{Y}_{HL}^{(n)},{Y}_{HH}^{(n)}\}).
\end{equation}


{\flushleft\textbf{Spiking PSC.}} Unlike spiking PWC, spiking PSC incorporates a random shuffle and inverse shuffle operation in the final-level wavelet domain to model long-range dependencies, which can be expressed as:
\begin{equation}
	Z^{(f)}=IDWT(\mathcal{IS}(SSepConv (\mathcal{RS}(\mathrm{DWT}({X}_{LL}^{(f)}))))),
\end{equation}
where $\mathcal{RS}$ and $\mathcal{IS}$ denote the random shuffle and inverse shuffle, respectively. $f$ means the wavelet domain at the final level. The high-frequency parts follow a similar pattern.

\begin{figure*}
	\centering
	\includegraphics[width=0.9 \linewidth]{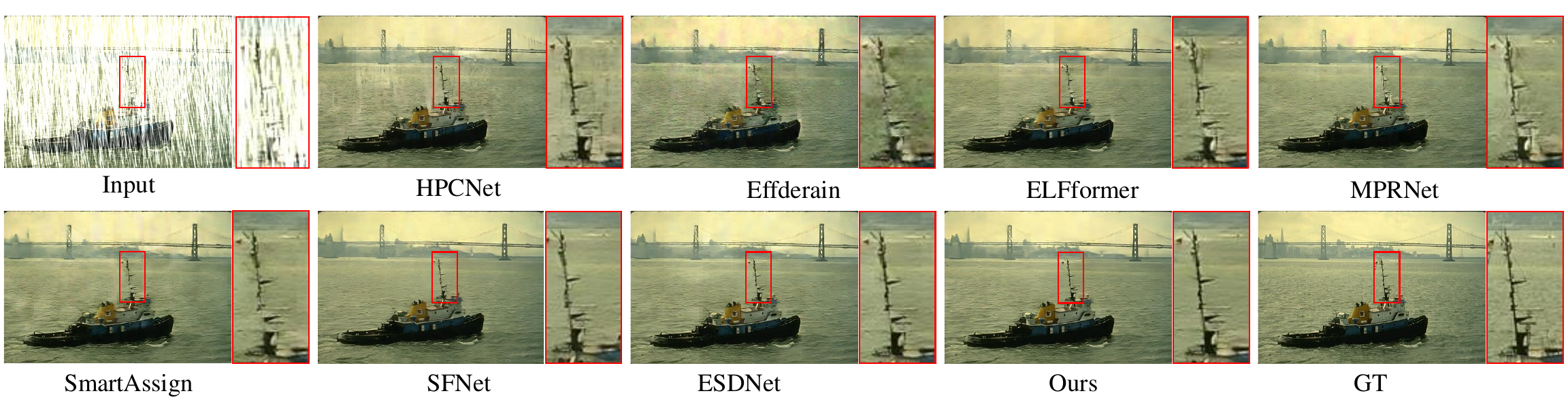}
	\vspace{-3mm}
	\caption{Visual comparison on the Rain200H dataset for image deraining task. Please zoom in to see the details.}
	\vspace{-3mm}
	\label{fig4}
\end{figure*}

\begin{figure*}
	\centering
	\includegraphics[width=0.9 \linewidth]{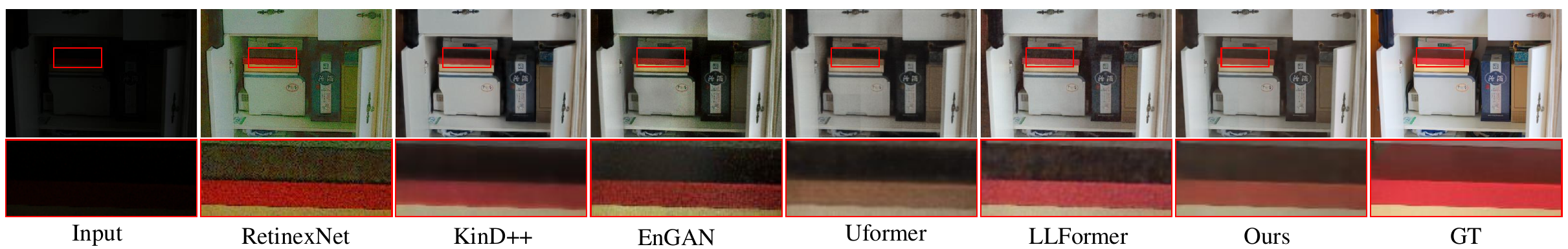}
	\vspace{-3mm}
	\caption{Visual comparison on the LOL dataset for image enhancement.  Please zoom in to see the details.}
	\vspace{-3mm}
	\label{fig6}
\end{figure*}
\vspace{-2mm}
\subsection{Multi-scale Progressive SCAM}
The conversion of discrete pulse sequences into continuous pixel values commonly  employs mean sampling over the time dimension, which leads to the loss of structural information \cite{song2024learning}. To address  this issue, we apply a multi-scale progressive fusion strategy to mitigate the missing of information. Moreover, we design an ultra-lightweight spatial-channel adjustment module (SCAM) to adaptively learn richer feature representations. In SCAM, we first project the input feature $\boldsymbol{F}$, and obtain: 
\begin{equation}
	\hat{\boldsymbol{F}},\boldsymbol{Q_c},\boldsymbol{Q_s}=\mathrm{Split}\left(\phi_1\left(\mathrm{Conv}(\boldsymbol{F})\right)\right),
\end{equation}
where $\phi_1$ indicates the GeLU function. $\hat{\boldsymbol{F}}$ represents the  projection of the input $\boldsymbol{F}$, and $\boldsymbol{Q_c}$ and $\boldsymbol{Q_s}$ aim to learn the attention matrix of channel and spatial dimension, respectively. Then, we can get the attention maps: 
\begin{equation}
	\hat{\boldsymbol{Q_c}}=\sigma\left(\mathrm{Conv}\left(\phi_2\left(\mathrm{Conv}(\mathrm{GAP}(\boldsymbol{Q_c}))\right)\right)\right),
\end{equation}
\begin{equation}
	\hat{\boldsymbol{Q_s}}=\mathrm{DConv}_{1\times k}^s(\mathrm{DConv}_{k\times 1}^s(\boldsymbol{Q_s})),
\end{equation}
where  $\operatorname{GAP}(\cdot)$ denotes the global average pooling, and $\sigma(\cdot)$ indicates the Sigmoid operation. $\mathrm{DConv}_{1\times k}^s$ and $\mathrm{DConv}_{k\times 1}^s$ refer to  striped depth-wise convolution with a large kernel. 

\begin{equation}
	\hat{\boldsymbol{F_{out}}}=(1-\alpha)\otimes \hat{\boldsymbol{Q_s}}\otimes \hat{\boldsymbol{F}}+\alpha\cdot \hat{\boldsymbol{Q_c}}\otimes \hat{\boldsymbol{F}},
\end{equation}
where $\alpha$ is a learnable parameter. 
Finally, the output feature $\hat{\boldsymbol{F_{out}}}$  may be fed into a $3 \times3$ convolution layer to obtain the output image of SPWM.
\vspace{-2mm}
\subsection{Network Training}
In recent years, backpropagation-based training schemes have achieved significant progress in facilitating the training of SNNs, showing considerable superiority over transformation-based strategies.
Since the non-differentiable nature of binary activations in SNN neurons, most existing methods use gradient-proxy functions to implement backpropagation training. 
We use the Sigmoid  as the gradient-surrogate function learning to train SNNs. The gradient surrogate function can be defined as:
%
%
\begin{equation}
	\begin{split}
		\sigma^{\prime}(x)=\delta \cdot &\frac{1}{1+e^{-\delta x}} \cdot(1-\frac{1}{1+e^{-\delta x}}),
	\end{split}
\end{equation}
where $\delta$ refers to a hyper-parameter, adopted to adjust the gradient of the surrogate function. The larger the value of $\delta$, the greater the gradient of the function.

We employ SSIM loss  and train the model by minimizing it.
The loss function can be depicted as:
\begin{equation}
	\mathcal{L}_{ssim}=1-\operatorname{SSIM}(\mathcal{SPWM}(\mathcal{X}), \mathcal{Y}),
\end{equation}
where $\mathcal{SPWM}(\cdot)$ is the proposed SPWM, $\mathcal{X}$ and $\mathcal{Y}$ are the input and its corresponding ground-truth, respectively.

\begin{table*}[!ht]
	\centering
	\caption{Comparison of quantitative results on three image deraining datasets. \textbf{Blod}  indicate the best  results.}
    
	\resizebox{0.65\width}{!}{
		\begin{tabular}{cc|cccccc|ccc}
			\hline
			\multicolumn{2}{c|}{} &                       \multicolumn{2}{c}{\textbf{Rain200L}} & \multicolumn{2}{c}{\textbf{Rain200H}}      & \multicolumn{2}{c|}{\textbf{Rain1200}}       &   &         &                                        \\
			\multicolumn{2}{c|}{\multirow{-2}{*}{\textbf{Model}}}                     & \textbf{PSNR$\uparrow$}               & \textbf{SSIM$\uparrow$}                & \textbf{PSNR$\uparrow$}               & \textbf{SSIM$\uparrow$}                & \textbf{PSNR$\uparrow$}               & \textbf{SSIM$\uparrow$}                                & \multirow{-2}{*}{\textbf{Params(M)$\downarrow$}} & \multirow{-2}{*}{\textbf{FLOPs(G)$\downarrow$}} & \multirow{-2}{*}{\textbf{Energy(uJ)$\downarrow$}} \\ \hline
			\multicolumn{2}{c|}{RCDNet \cite{wang2020RCDNet} }                       & 38.84                        & 0.9854                        & 28.98                        & 0.8889                        & 32.68                        & 0.9189                                             & 2.958                                & 194.501                             & 2.431$\times10^6$            \\
			\multicolumn{2}{c|}{ MPRNet \cite{zamir2021multi} }    
			&39.82 &0.9863
			&29.94  & 0.8999  & 34.50    & 0.9369            
			& 3.637  & 548.651   & 6.858$\times10^6$   \\
			\multicolumn{2}{c|}{  Effderain \cite{guo2021efficientderain}  }   & 36.06   & 0.9731 
			& 26.11   & 0.8341      & 32.85        & 0.9147     & 27.654        & 52.915 & 6.614$\times10^5$            \\
			\multicolumn{2}{c|}{  SFNet  \cite{cui2022SFNet}  }            & 39.50   & 0.9850    & 29.75    &0.9008             &34.51 &0.9383
			& 13.234    & 124.439     & 1.555$\times10^6$     
			\\ 
			\multicolumn{2}{c|}{ DRT \cite{liang2022drt}   }    & 38.81  & 0.9830  & 28.67                   & 0.8796   & 33.88   & 0.9283  & 1.176  & 166.045                  & 2.075$\times10^6$            \\
			\multicolumn{2}{c|}{ ELFformer \cite{jiang2022elformer}  }   & 38.85  & 0.9800              & 28.93   & 0.8852  & 33.54   & 0.9360     & 1.221   & 23.667  & 2.958$\times10^5$            
			\\
			\multicolumn{2}{c|}{HPCNet \cite{wang2023hpcnet}}         & 39.14               & 0.9847    & 29.17   &0.8962        & 34.46       & 0.9378       & 1.411   & 29.540       & 3.692$\times10^5$            	\\
			\multicolumn{2}{c|}{ SmartAssign   \cite{wang2023smartassign}  }    & 38.41    & 0.9814   & 27.71   & 0.8536    & 33.11                    
			& 0.9154       & 1.359   & 90.386    & 1.129$\times10^6$            \\ 
			\multicolumn{2}{c|}{  ESDNet   \cite{song2024learning}  }   & 39.85 &0.9869 & 29.50  & 0.9079 & 34.25 &0.9367 &0.165 &7.320  & 9.150$\times 10^4$
			\\ \hline
			\multicolumn{2}{c|}{  Ours    }      & \textbf{40.58} & \textbf{0.9884} & \textbf{30.49} & \textbf{0.9217 } & \textbf{34.66} &\textbf{0.9392} &\textbf{0.125} &\textbf{2.912}  & \textbf{3.640$\times 10^4$} 
			\\ \hline           
		\end{tabular}
	}
	
	\label{tab3}
\end{table*}

\begin{table}[ht]
	\centering
	\caption{Comparison of quantitative results on  the LOL dataset for low light image enhancement.
	}
	\resizebox{0.70\width}{!}{
		\begin{tabular}{c|ccccc}
			\toprule
			\multirow{1}{*}{\textbf{Model}}  & \multirow{1}{*}{\textbf{PSNR}$\uparrow$} & \multirow{1}{*}{\textbf{SSIM}$\uparrow$}  & \multirow{1}{*}{\textbf{LPIPS}$\downarrow$} & \multirow{1}{*}{\textbf{FLOPs(G)}$\downarrow$}  & \multirow{1}{*}{\textbf{Params(M)}$\downarrow$} \\ \hline
			RetinexNet \cite{wei2018deep}                 & 16.77 & 0.425                                  &  0.4739     & 148.54  &0.84               \\
			DRBN      \cite{yang2020fidelity}         & 15.15 &0.492 &0.339    &  42.41 &0.58           \\
			EnGAN         \cite{jiang2021enlightengan}         & 17.48         & 0.652                               &  0.322       &72.61         &8.37                 \\
			
			GLADNet        \cite{wang2018gladnet}              & 19.72             &0.682                                          & 0.321  & 275.32 &1.13                \\
			KinD     \cite{zhang2019kindling}                  &   17.64& 0.771& 0.175      & 36.57  & 8.54                                               \\
			KinD++     \cite{zhang2021beyond}                  &21.80              &0.829                                                & 0.158 & 2970.50 & 8.28           \\
			Uformer     \cite{wang2022uformer}              &18.54 &0.721& 0.320 & 45.9  & 50.88          \\
			LLFormer    \cite{wang2023ultra}            &\textbf{23.65 }             &0.816                      & 0.169 &22.034 &   24.51         \\  
			CDAN     \cite{shakibania2025cdan}           &20.10              &0.816                      & 0.167 &8.41 &   3.58        \\	\hline
			Ours                      &     22.35            &           \textbf{ 0.844  }                      &\textbf{0.155}             &\textbf{2.912} &                   \textbf{0.125}                   \\
			\bottomrule
		\end{tabular}
	}
	\label{tab4}
\end{table}

\section{Experiments}
We conduct comprehensive experiments on  IR tasks to evaluate the effectiveness of the proposed SPWM, including  image deraining and
image enhancement.  

{\flushleft\textbf{Implementation Details.}} 
Our network, implemented using PyTorch 2.3, underwent training and testing on an NVIDIA A6000 GPU.
We set the learning rate to $1\times 10^{-3}$ and apply the cosine annealing strategy to steadily decrease the final learning rate to $1 \times10^{-7}$. In the testing phase, we  apply the sliding window slicing strategy for testing. For the $\delta$ of the gradient proxy function, it is set to 4 according to EMS-YOLO \cite{su2023deep}. 
For image deraining and denoising tasks, to ensure a fair comparison with the SNN-based methods\cite{song2024learning}, we conduct experiments with an Adam optimizer, a patch size of 64, a batch size of 12, and a time step of 4. Moreover, we retrained the baselines under same settings on our device and environment. For the image enhancement task, we set the patch size to 128 and the batch size to 6, and the total number of training epochs is 1000. 

{\flushleft\textbf{Evaluation Metrics}.} We mainly adopt Peak Signal to Noise Ratio (PSNR) and Structural Similarity (SSIM)  to evaluate the performance of networks.  For image deraining tasks, following the previous methods in the \cite{song2024learning}, PSNR/SSIM scores are calculated on the Y channel of the YCbCr space. Additionally,
LPIPS   are utilized to evaluate perceptual performance.  

{\flushleft\textbf{Energy Consumption}.} In our work, ANN part (the first conv, SCAM, and last conv) require FLOPs. The other layers are calculated through SOPs. The sign function is  in LIFs, which have fewer number than the conv layers. Following \cite{song2024learning}, we calculate the energy as follows: a FLOP requires 12.5 pJ, a SOP 77 fJ, and a Sign 3.7 pJ.

\begin{table}[ht]
	\centering
	\caption{Quantitative results of different noise levels \(\sigma\) on the CBSD68 dataset for gaussian denoising.}
	\resizebox{0.75\width}{!}{
		\begin{tabular}{c | c c c c c c c c}
			\toprule
			\multirow{2}{*}{\textbf{Methods}} 
			& \multicolumn{2}{c}{\(\sigma = 15\)} 
			& \multicolumn{2}{c}{\(\sigma = 25\)} 
			& \multicolumn{2}{c}{\(\sigma = 50\)} \\
			
			\cmidrule(lr){2-3}\cmidrule(lr){4-5}\cmidrule(lr){6-7}
			&	PSNR & SSIM & PSNR & SSIM & PSNR &SSIM &Params & FLOPs \\
			\midrule

			MSANet  \cite{gou2022multi}  & 33.10 & 0.9143 & 30.67  &0.8693  & 27.05 & 0.7725    & 7.997 &35.35 \\
			DeamNet \cite{ren2021adaptive}  & 33.43 &0.9295  &30.57  &0.8832  &27.06  &  \textbf{0.7943}  &1.876&145.8 \\
			DRANet \cite{wu2024dual}  & 33.18 & 0.9244 & 30.39 &0.8774 & 27.16 & 0.7896 & 1.618&	592.33\\
			Xformer \cite{zhang2023xformer}  & 33.47 & 0.9203 & 30.53 & 0.8586 &   \textbf{27.45} & 0.7852 &25.12&	143.08\\
			\midrule
			Ours     &   \textbf{33.67} &   \textbf{0.9309} &  \textbf{ 30.95 }&   \textbf{0.8861} & 27.42 & 0.7936  &  \textbf{0.125}&	  \textbf{2.912} \\
			\bottomrule
	\end{tabular}  }

	\label{tab:denoise}
\end{table}
\subsection{Image Deraining}

We perform the deraining experiments on Rain200L \cite{yang2017JORDER}, Rain200H \cite{yang2017JORDER} and Rain1200 \cite{zhang2018diddata}.  
Table~\ref{tab3} provides a comprehensive comparison between our method and 9 competitive baselines, including four CNN-based methods (RCDNet~\cite{wang2020RCDNet}, MPRNet~\cite{zamir2021multi}, Effderain~\cite{guo2021efficientderain}, SFNet~\cite{cui2022SFNet}), four Transformer-based methods (DRT~\cite{liang2022drt}, ELFormer ~\cite{jiang2022elformer}, HPCNet \cite{wang2023hpcnet} and SmartAssign ~\cite{wang2023smartassign}) and a SNN-based method ( ESDNet \cite{song2024learning}). Our SPWM clearly boosts performance, enhancing PSNR and SSIM metrics, outpacing all baselines. Notably, on Rain200H, our technique surpasses ESDNet, a SNN-based method, by 0.99 dB in PSNR. This superior outcome illustrates a fresh perspective for SNN architectures in IR tasks. Furthermore, Fig.~\ref{fig4}  provides qualitative evaluations on Rain200H. The visual comparisons demonstrate heightened contrast and diminished color distortion in our method, aligning with quantitative data. 
Our model retains finer details and achieves higher perceptual quality.

\subsection{Image Enhancement}
The proposed SPWM is evaluated for low light image
enhancement task on LOL dataset \cite{wei2018deep}. LOL dataset consists of 500 low and normal light image pairs, and we splits 485 for training and 15 for testing. 
We compare our spiking paradigm with ten ANN low-light
image enhancement methods, including  RetinexNet\cite{wei2018deep},   DRBN\cite{yang2020fidelity}, EnGAN\cite{jiang2021enlightengan},  GLADNet\cite{wang2018gladnet}, 	KinD\cite{zhang2019kindling},   	KinD++\cite{zhang2021beyond}, Uformer\cite{wang2022uformer},  LLFormer\cite{wang2023ultra}, CDAN\cite{shakibania2025cdan}.  Table~\ref{tab4} presents the quantitative results of our method. SPWM achieves the highest SSIM scores and very competitive PSNR values, while maintaining the lowest parameter and FLOPs. Additionally, our method excels in perceptual quality metrics. Fig.~\ref{fig6} illustrates the visual comparisons, where our approach significantly outperforms other methods in terms of detail preservation and overall quality.

\begin{table}[ht]
\centering
\caption{Comparison for image dehazing on the Dense-Haze dataset \cite{ancuti2019dense}.}
\label{tab:haze}
\begin{tabular}{l|ccc}
\hline
\textbf{Method} & \textbf{PSNR $\uparrow$} & \textbf{SSIM $\uparrow$} & \textbf{Params $\downarrow$} \\ \hline
FFA-Net \cite{qin2020ffa} & 14.39 & 0.4524 & 4.68M \\
MSBDN \cite{dong2020multi} & 15.37 & 0.4858 & 31.35M \\
Fourmer \cite{zhou2023fourmer} & \textbf{15.95} & 0.4917 & 1.29M \\ \hline
\textbf{Ours} & 15.70 & \textbf{0.5445} & \textbf{0.125M} \\ \hline
\end{tabular}
\end{table}
\subsection{Image Dehazing}
 For image dehazing, we perform the dehazing experiments on Dense-Haze \cite{ancuti2019dense}. Table~\ref{tab:haze} presents the quantitative results for image dehazing. We compare our spiking paradigm with three dehazing methods, including FFA-Net \cite{qin2020ffa}, MSBDN \cite{dong2020multi}, and Fourmer \cite{zhou2023fourmer}. SPWM achieves the highest SSIM scores and highly competitive PSNR values, while maintaining the lowest parameter count and FLOPs. The slight variation in PSNR aligns with our observation that global degradations (haze) pose a different challenge than local ones (rain).

\begin{table}[ht]
\centering
	\caption{Effect of different time steps.}
    \resizebox{0.65\linewidth}{!}{
      \begin{tabular}{ccccc}
        \specialrule{1.2pt}{0.2pt}{1pt}
        Time steps & FLOPs (G) & Params (K) & PSNR & SSIM \\
        \hline
        1 & 2.7852 & 125.21 & 29.91 & 0.9116 \\
        2 & 2.8277 & 125.23 & 30.08 & 0.9138 \\
        3 & 2.8702 & 125.25 & 30.38 & 0.9183 \\
        4 & 2.9126 & 125.28 & 30.49 & 0.9217 \\
        5 & 2.9551 & 125.30 & 30.60 & 0.9223 \\
        \specialrule{1.2pt}{0.2pt}{1pt}
      \end{tabular}
    }
    \label{tab5}
\end{table}

\begin{table}[ht]
\centering
	\caption{Effect of different DWT levels.}
    \resizebox{0.65\linewidth}{!}{
      \begin{tabular}{ccccc}
        \specialrule{1.2pt}{0.2pt}{1pt}
        DWT level & FLOPs (G) & Params (K) & PSNR & SSIM \\
        \hline
        0 & 2.9126 & 125.28 & 29.59 & 0.9103 \\
        1 & 2.9126 & 125.28 & 30.15 & 0.9153 \\
        2 & 2.9126 & 125.28 & 30.37 & 0.9195 \\
        3 & 2.9126 & 125.28 & 30.26 & 0.9209 \\
        Our set & 2.9126 & 125.28 & 30.49 & 0.9217 \\
        \specialrule{1.2pt}{0.2pt}{1pt}
      \end{tabular}
    }
    \label{tab6}
\end{table}

\begin{table}[ht]
\centering
	\caption{Ablation studies for different network architecture.}
    
	\resizebox{0.850\width}{!}{
		\begin{tabular}{cccccccc}
			\specialrule{1.2pt}{0.2pt}{1pt}
			Model & SPWU &SPSU & MDA                 & MSPS & SCAM                  & PSNR   & SSIM   \\ \hline
			A   &$\times$   &           $ \checkmark$        &   $ \checkmark$  & $ \checkmark$    &      $ \checkmark$             & 30.28 & 0.9168   \\
			B   & $ \checkmark$   &    $\times$              &  $ \checkmark$    &  $ \checkmark$    &      $ \checkmark$              & 30.18 & 0.9159     \\ \hline
			C   & $\checkmark$  & $ \checkmark$ &$\times$  &  $\times$  & $\times$& 29.05 &0.9000  \\ 
			D   & $\checkmark$  & $ \checkmark$ &$\checkmark$   & $\times$      &$\times$    &29.48     &  0.9050\\
			\hline
			E   &   $ \checkmark$  & $\checkmark$                  &  $ \checkmark$   &  $\times$    & $\checkmark$                   & 30.27 & 0.9178 \\
			F   &   $ \checkmark$  & $ \checkmark$                &  $ \checkmark$     & $ \checkmark$   &        $\times$               &30.12    & 0.9164  \\ \hline
			G   &   $ \checkmark$   & $ \checkmark$               &     $ \checkmark$    & $ \checkmark$   & $ \checkmark$                    &30.49    &0.9217       \\ \specialrule{1.2pt}{0.2pt}{1pt}
		\end{tabular}
	}
	
	\label{tab7}
\end{table}

\vspace{-3mm}
\subsection{Ablation Study and Analysis}
We conduct extensive ablation experiments to validate the effectiveness of each contribution, 
 providing a deeper understanding of the role of each component.  Note that, all ablation experiments are conducted on the Rain200H dataset.

{\flushleft{\textbf{ Effect of different time steps}}.} The ablation results  with different time steps are shown in Table~\ref{tab5}.
Based on the quantitative results, it becomes evident that as the time steps increase, the quality of image restoration improves, which suggests that extended time steps enable the model to more effectively capture richer representations. However, this comes with a notable rise in computational cost. To strike an optimal balance between performance and efficiency, we opted for a time step of 4 as the standard setting in our study.

\begin{figure}[!t]
	\centering
	\includegraphics[width=1\linewidth]{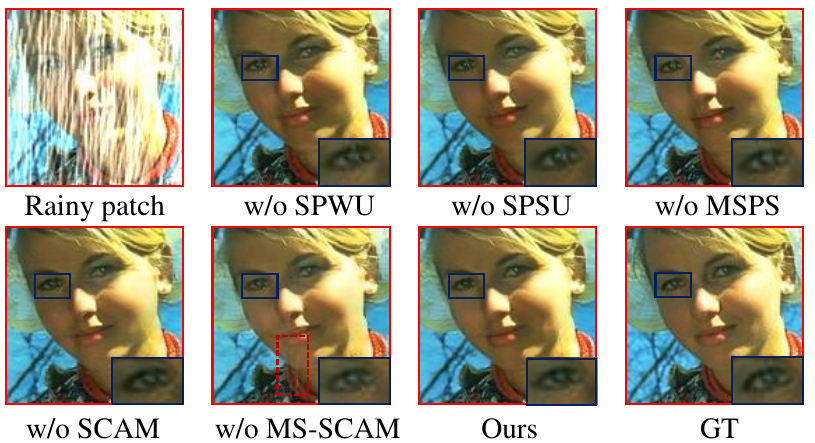}
	\caption{Visual comparison on the ablation study.}
	\label{fig8}
\end{figure}

\begin{figure}[!t]
	\centering
	\includegraphics[width=1 \linewidth]{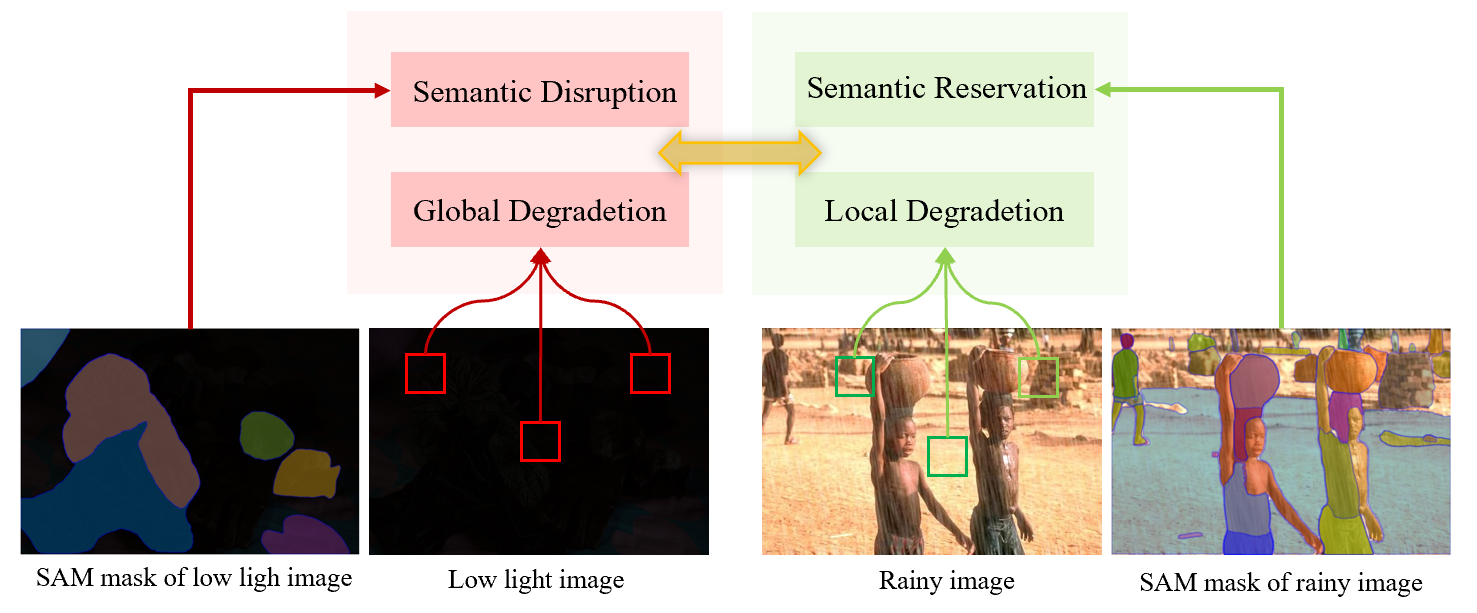}
	\caption{Degradation comparison of different IR tasks.}
	\label{fig7}
\end{figure}

{\flushleft{\textbf{Effect of different DWT levels }}.} 
Table~\ref{tab6} illustrates the impact of DWT levels on the model's performance. The 1-level DWT-based model significantly outperforms the model without DWT (0-level), suggesting a substantial improvement in performance by using DWT. Moreover, this improvement does not incur any additional parameter or FLOPs overhead. 
SPWM employs 2-level wavelet transform in the downsampling and upsampling blocks and 1-level wavelet transform in the  bottleneck blocks. This setup achieves superior performance, because after downsample layers, the feature sizes are already relatively small, making it challenging for higher levels of DWT to extract meaningful information.

{\flushleft{\textbf{Effect of different network architecture}.}}
Table~\ref{tab7} demonstrates the impact of all components on the network's performance. Fig.~\ref{fig8} presents the  visual comparison results. 
MSPS means the multi-scale progressive fusion strategy, and MS-SCAM refers to the  MSPS and SCAM.  The inferior performance of A and B  models compared to G substantiates the value of the SPWU and SPSU within the overall architecture. The significantly reduced performance of models C and D suggests that the MAU, MSPS, and SCAM effectively mitigate the information loss inherent in SNNs. \textit{This also implies that effectively combining SNN and ANN can achieve higher performance at a lower cost, offering new perspectives for IR technologies in resource-constrained scenarios.} Finally, model G achieves the best performance by using all technologies, proving that each part of SPWM is useful for IR tasks.

\begin{table}[ht]
\centering
\caption{Efficiency and performance analysis under different time steps $T$ on the Rain200H dataset. The inference time is measured in seconds (s).}
\label{tab:time}
\begin{tabular}{l|ccccc|c}
\hline
\textbf{Method} & \multicolumn{5}{c|}{\textbf{Ours}} & \textbf{ESDNet \cite{song2024learning}} \\ \hline
Time steps ($T$) & 1 & 2 & 3 & 4 & 5 & 4 \\ 
Inference time (s) & 1.72 & 1.90 & 2.13 & 2.46 & 2.59 & 2.98 \\ 
PSNR (dB) & 29.91 & 30.08 & 30.38 & 30.49 & 30.60 & 29.50 \\ \hline
\end{tabular}
\end{table}
{\flushleft{\textbf{Efficiency analysis under different time steps}.}}
As shown in Table \ref{tab:time}, we report the inference time and restoration performance (PSNR) of our model across various time steps $T \in \{1, 2, 3, 4, 5\}$. It is important to note that while current GPU and CPU architectures are heavily optimized for Artificial Neural Networks (ANNs) rather than the asynchronous, event-driven nature of SNNs, our model still demonstrates superior efficiency.  As $T$ increases, the PSNR consistently improves, allowing users to balance restoration quality and computational cost depending on the application scenario. Compared to the recent SNN-based method ESDNet, our model achieves a significantly higher PSNR (30.49 dB vs. 29.50 dB) even with the same number of time steps ($T=4$), while being approximately 17.4\% faster.

\begin{table}[ht]
\centering
\caption{Ablation study of the Random Shuffling (RS) operation and comparison with state-of-the-art SNN-based methods on the Rain200H dataset.}
\label{tab:Shuffling}
\begin{tabular}{l|cccc}
\hline
\textbf{Method} & \textbf{PSNR (dB)} & \textbf{Params (M)} & \textbf{FLOPs (G)} & \textbf{Inference Time (s)} \\ \hline
w/o RS          & 30.28             & 0.125               & 2.91               & 2.45                        \\ 
w/ RS    & \textbf{30.49}    & 0.125               & 2.91               & 2.46                        \\ \hline
ESDNet \cite{song2024learning} & 29.50             & 0.165               & 7.32               & 2.98                        \\ \hline
\end{tabular}
\end{table}

{\flushleft{\textbf{Effect of random shuffling}.}}
As shown in Table \ref{tab:Shuffling}, we conduct a comparative analysis of the model with and without the RS operation. The inclusion of RS improves the PSNR from 30.28 dB to 30.49 dB (+0.21 dB). Crucially, this gain is achieved with zero additional parameters and zero additional FLOPs, as RS is a non-parametric indexing operation. The negligible increase in inference time (0.01s) is merely due to memory access overhead on current hardware, not computational redundancy. By disrupting the spatial locality, RS allows the subsequent convolutional kernels to aggregate information from distant wavelet coefficients. This acts as a ``pseudo-global'' attention mechanism that is significantly more cost-effective than standard self-attention or large-kernel convolutions. Compared to the recent SNN-based method ESDNet \cite{song2024learning}, our model with RS achieves substantially higher performance (30.49 dB vs. 29.50 dB) while maintaining significantly lower parameter count (0.125M vs. 0.165M) and computational cost.

{\flushleft{\textbf{Analysis on different IR tasks}.}}  Our SPWM achieves a PSNR that is 1.3 dB lower than LLFormer for image enhancement, yet shows a significant improvement in image deraining. This raises the question: \textit{Why do SNN-based model perform less favorably in enhancement tasks compared to deraining tasks?}
Fig.~\ref{fig7} reveals the distinct characteristics of low-light images compared to rainy images, specifically \textit{severe semantic structure disruption and global degradation}. The two characteristics significantly exacerbate the issue of information loss due to the binary spike signals in SNNs. This observation aligns with our denoising experiments, where larger noise levels (greater degradation) typically result in weaker performance. Ideas to alleviate the problem mainly include the use of advanced ANN techniques and the introduction of additional  prior information in SNNs, which is also our future research direction.

\vspace{-10pt}

\section{Conclusion}
This paper presents a spiking pyramid wavelet-based model (SPWM) for image restoration (IR). SPWM introduces a novel wavelet-based SNN paradigm,
which explicitly model long-range dependency and exploit the properties of the degradation in the wavelet domain.  Extensive experiments demonstrate that our proposed method achieves competitive performance with low resource consumption. Our method highlights the potential of SNNs for IR  and offers novel perspectives for future applications on resource-limited devices. Finally, in ablation studies, we reveal the challenges faced by SNNs  and suggest directions for future development.
\section{Acknowledgments}
 This work was supported by Natural Science Foundation of China: No. 62406135, Natural Science Foundation of Jiangsu Province: BK20241198,
Gusu Innovation and Entrepreneur Leading Talents: No. ZXL2024362 and Nanjing University-China Mobile Communications Group Co. Ltd. Joint Institute.

\bibliographystyle{elsarticle-num} 
\bibliography{reference}

\end{document}